\pdfoutput=1

\documentclass[letterpaper]{article} 

\usepackage[preprint]{aaai2027}

\usepackage[hyphens]{url}  
\usepackage{graphicx} 
\usepackage{natbib}  
\usepackage{caption} 
\usepackage{algorithm}
\usepackage{algorithmic}
\usepackage{amsmath}
\usepackage{amssymb}
\usepackage{booktabs}
\usepackage{multirow}

\graphicspath{{figures/}}

\newcommand{\best}[1]{\textbf{#1}}
\newcommand{\second}[1]{\underline{#1}}
\newcommand{\ci}[2]{#1\,\ensuremath{{\scriptstyle\pm\,#2}}}

\title{STRATA: Self-Maintaining Structured Experience Graphs\\for Self-Evolving Language Agents}

\author{
    Guanhua Zhu\textsuperscript{\rm 1},
    Hanwen Li\textsuperscript{\rm 1},
    Zhi Wei\textsuperscript{\rm 2},
    Mengnan Du\textsuperscript{\rm 3,*},
    Chenxi Yuan\textsuperscript{\rm 1,*}
}
\affiliations{
    \textsuperscript{\rm 1}Department of Informatics, New Jersey Institute of Technology\\
    \textsuperscript{\rm 2}Department of Computer Science, New Jersey Institute of Technology\\
    \textsuperscript{\rm 3}School of Artificial Intelligence, The Chinese University of Hong Kong, Shenzhen\\
    gz72@njit.edu, hl544@njit.edu, zhi.wei@njit.edu, mengnandu@cuhk.edu.cn, chenxi.yuan@njit.edu\\
    \textsuperscript{*}Corresponding authors
}

\begin{document}
\maketitle

\begin{abstract}
Large language model (LLM) agents can \emph{self-evolve} at test time by
accumulating experience: retrieving relevant memories before acting and writing
new lessons back afterwards. Yet most are \emph{single-level}: trajectories,
distilled workflows and reasoning items all land in one undifferentiated pool
indexed by embedding similarity, with no typed abstraction hierarchy. They never
forget or deduplicate, and are prone to \emph{negative transfer}, where a
superficially similar but directionally opposite experience is recalled and its
concrete steps are copied verbatim, actively harming the agent. We introduce
STRATA, a structured, outcome-grounded, self-maintaining memory that
organizes experience into a three-layer graph of atomic skills and code
snippets, task workflows and generalized meta-frameworks, connected by typed
edges.
STRATA maintains itself through five mechanisms sharing one similarity geometry:
a multi-dimensional similarity that scores retrieval by operation and target; a
typed reflector crystallizing procedures from successes and pitfalls from
failures; community detection that induces meta-frameworks; pressure-triggered
garbage collection with articulation-point protection; and \emph{contrast edges}
that flag confusable, direction-opposite experiences. Across four interactive
benchmarks (AppWorld, WebArena, ALFWorld, ScienceWorld), two frozen backbones and
seven baselines, STRATA attains the best average score while simultaneously using the lowest latency and fewest environment steps of
any memory-augmented method, and 44\% fewer lines of injected memory than the
most context-hungry baseline. We isolate two failure modes of prior memories: single-level memory
degrades sharply as the task set grows, and single-level concrete memories suffer
catastrophic negative transfer, which contrast edges repair, lifting
ScienceWorld from $78.7$ to $96.7$ within the same system.
\end{abstract}

\section{Introduction}

Language-model agents increasingly solve long-horizon interactive tasks such as
navigating websites, orchestrating APIs across applications, and acting in
simulated environments~\citep{yao2022react,trivedi2024appworld,zhou2024webarena,shridhar2020alfworld,wang2022scienceworld},
building on prompting and tool-use techniques that structure a single
episode~\citep{wei2022chain,yao2023tree,schick2023toolformer,wang2024survey}.
A promising route to improving such agents without weight updates is
\emph{experience-driven self-evolution}: the agent keeps a memory of its past
episodes, retrieves relevant experience before each new task, and folds fresh
lessons back into memory afterwards, becoming more capable over
time~\citep{ouyang2025reasoningbank,xu2026mem,shinn2023reflexion}. Because this loop runs
purely at test time, its effectiveness hinges almost entirely on how experience is represented, retrieved and maintained~\citep{snell2024scaling}.

\begin{figure}[t]
\centering
\includegraphics[width=\columnwidth]{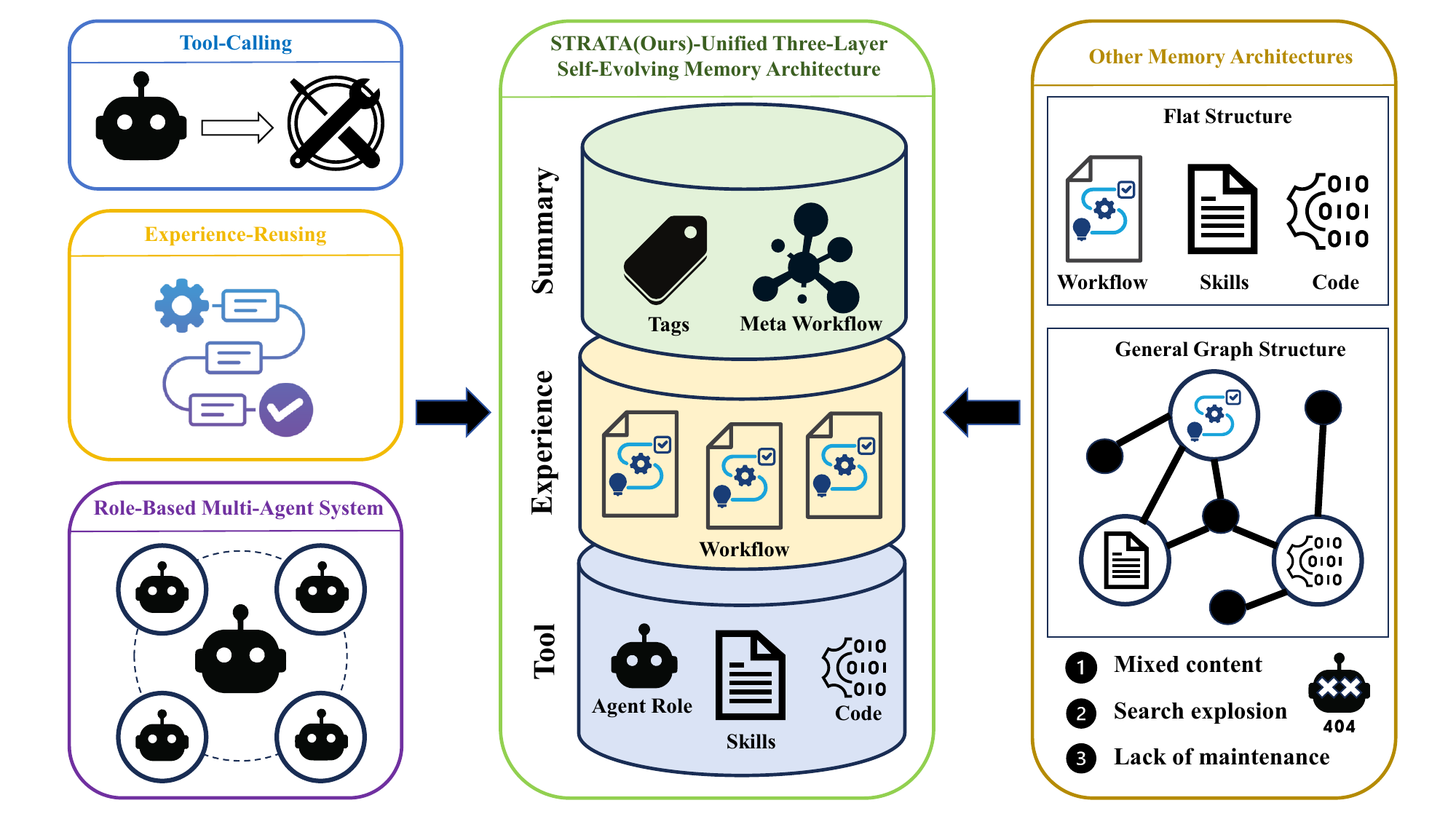}
\caption{Agent memories are asked to serve tool calling, experience reuse and
role-based decomposition at once (left). Existing memories are either flat lists or generic graphs, which mix content of different
granularity, dilute retrieval as they grow, and are never maintained (right).
STRATA unifies the three in a single three-layer self-evolving graph (middle).}
\label{fig:intro}
\end{figure}

Despite rapid progress, prevailing agent memories share three limitations.

\emph{(1) They lack a typed abstraction hierarchy.} Several do organize what they
store (AWM distils workflows from trajectories~\citep{wang2024agent}, A-Mem links
notes into a network~\citep{xu2026mem}), but within one undifferentiated level:
nothing separates a one-off tool call from a reusable procedure from a cross-task
principle, and no operation folds procedures into a principle. Retrieval stays a
single similarity search dominated by surface overlap, merging same-material, different-operation pairs like ``create a playlist'' and ``add months
to a note''.

\emph{(2) They are not self-maintaining.} Memory grows monotonically, with no
forgetting, no deduplication and no consolidation, so stale entries crowd out
useful ones. We measure this: with forgetting disabled the workflow store grows
to $269$ nodes injecting $48.7$ lines of memory per task, yet scores lower than a bounded $201$-node memory injecting less than half as much.

\emph{(3) They are vulnerable to negative transfer.} When two experiences are
superficially similar but directionally opposite, as in ``heat the
substance'' versus ``cool the substance'', similarity recall surfaces the wrong
one, and if the memory stores concrete executable steps the agent copies them and
acts catastrophically. This is not hypothetical: on ScienceWorld a production
agent's native flat memory scores $-100$ on a task family that a memoryless agent
solves outright.

We address these limitations with STRATA, a memory that is structured, outcome-grounded and self-maintaining. STRATA
organizes experience into a three-layer directed graph: \emph{atomic} nodes
(reusable skills, code snippets, or specialized sub-roles), \emph{workflow}
nodes (generalized task procedures), and \emph{meta-framework} nodes (task
families abstracted from recurring workflows), linked by typed call,
generalization, merge and contrast edges. The graph is grown and pruned by five
mechanisms that all operate in one similarity geometry: \emph{(i)} a
multi-dimensional similarity combining content and goal embeddings with an
explicit operation/target tag term; \emph{(ii)} a typed reflector that
crystallizes a reusable procedure only from solved episodes and attaches
a pitfall bullet on failure, with a per-group adaptive threshold deciding
whether to create, extend, or skip a workflow; \emph{(iii)} community detection
that induces a meta-framework whenever a community is large, tight, and novel;
\emph{(iv)} pressure-triggered garbage collection that ranks workflows by a
competition-aware quality score and forgets the lowest while protecting
articulation points of the similarity graph; and \emph{(v)} contrast edges that connect confusable, direction-opposite experiences so that
retrieving one injects an explicit ``do not confuse'' warning about the other.

Our contributions are as follows.
\begin{itemize}\itemsep2pt
\item We propose STRATA, a three-layer structured experience graph for
self-evolving LLM agents, together with five maintenance mechanisms that grow,
generalize, and forget it, formulated precisely and reproducibly.
\item We evaluate STRATA on four interactive benchmarks across two frozen
backbones, seven baselines and three seeds, with a component-wise ablation of
every mechanism. STRATA obtains the best average score at the lowest latency and
fewest environment steps of any memory baseline.
\item We identify two failure modes of prior memories: a scale failure of
single-level memory that neither abstracts nor forgets (on Flash the strongest
baseline ranks first at $168$ tasks and seventh at $812$), and negative transfer in single-level concrete memories, which contrast edges repair, exposing a structure-versus-abstraction trade-off.
\end{itemize}

\section{Related Work}

\emph{Memory for LLM agents.}
A growing line of work equips LLM agents with an external memory that persists
across episodes. Early systems store and summarize raw interaction
history~\citep{packer2023memgpt,chhikara2025mem0,zhong2024memorybank}; Agent Workflow Memory
extracts reusable action workflows from past trajectories~\citep{wang2024agent};
Agentic Context Engineering (ACE) maintains an evolving, per-application
playbook of bullet-point strategies~\citep{zhang2025agentic}; ReasoningBank distills generalizable reasoning strategies from both successful and failed
trajectories~\citep{ouyang2025reasoningbank}. A-Mem organizes memories in a
Zettelkasten-style linked note network~\citep{xu2026mem}, MemGen weaves a generative latent memory into the agent's own reasoning~\citep{zhang2025memgen},
MemEvolve meta-evolves the memory architecture itself~\citep{zhang2025memevolve}, and
MemInsight autonomously augments stored items~\citep{salama2025meminsight}. These methods differ in what they store but share a common structure: a single-level collection retrieved by similarity, with no typed hierarchy
and, typically, no explicit forgetting or anti-confusion mechanism. A-Mem is the
most structured of them and, with MemEvolve, our strongest baseline: it links
notes into a network and lets new memories revise old ones, but the network is
single-level, one note per memory linked to its neighbours, so it neither
abstracts a family of notes into a reusable principle nor removes any note once
written. We show that this produces a sharp scale dependence. STRATA departs
from this line by making the memory a \emph{typed, layered graph} that is
actively maintained, consolidated into meta-frameworks, garbage-collected under
pressure and wired with contrast edges against negative transfer.

\emph{Self-evolving and experience-based agents.}
Reflexion converts episodic failures into verbal
self-feedback~\citep{shinn2023reflexion} and Self-Refine iterates on its own
output within an episode~\citep{madaan2023self};
Voyager accumulates a skill library of executable code~\citep{wang2023voyager};
ExpeL gathers cross-task insights for in-context reuse~\citep{zhao2024expel}; Synapse
retrieves trajectory exemplars~\citep{zheng2024synapse}; CoALA frames such designs as
cognitive architectures~\citep{sumers2023cognitive}, alongside simulacra systems
that persist state across episodes~\citep{park2023generative}. STRATA is complementary and
mechanism-level: rather than a new learning signal, it contributes the \emph{data
structure and maintenance policy} governing how experiences are typed,
connected, generalized and forgotten, so accumulated experience stays organized
as it scales.

\emph{Structured retrieval, forgetting, and negative transfer.}
Outside the agent setting, graph-structured and hierarchical
memories~\citep{edge2024local,gutierrez2024hipporag,gutierrez2025rag} improve retrieval
precision over flat retrieve-then-generate~\citep{lewis2020retrieval,karpukhin2020dense,izacard2021unsupervised}
and its refinements~\citep{gao2023precise,asai2024self}, while unstructured stores
suffer the classical catastrophic-forgetting and negative-transfer
problems~\citep{mccloskey1989catastrophic,kirkpatrick2017overcoming,parisi2018continual,zhang2022survey}.
Our maintenance machinery borrows standard graph tools: Louvain community
detection~\citep{blondel2008fast,newman2003finding}, conductance as a cut
quality measure~\citep{shi2000normalized}, and Tarjan's articulation-point
algorithm~\citep{tarjan1972depth}; the three-layer split mirrors the
episodic/semantic distinction and complementary learning
systems~\citep{tulving1972episodic,mcclelland1995there}. In the agent setting the
risk is sharper: because retrieved experience is \emph{executed}, an incorrectly
recalled procedure does not merely mislead but actively harms. STRATA adds an
explicit operation/target gate and surfaces the opposing experience as a
retrieval-time warning, a mechanism we are not aware of in prior agent memory.

\section{Method}
\label{sec:method}

\begin{figure*}[t]
\centering
\includegraphics[width=0.72\textwidth]{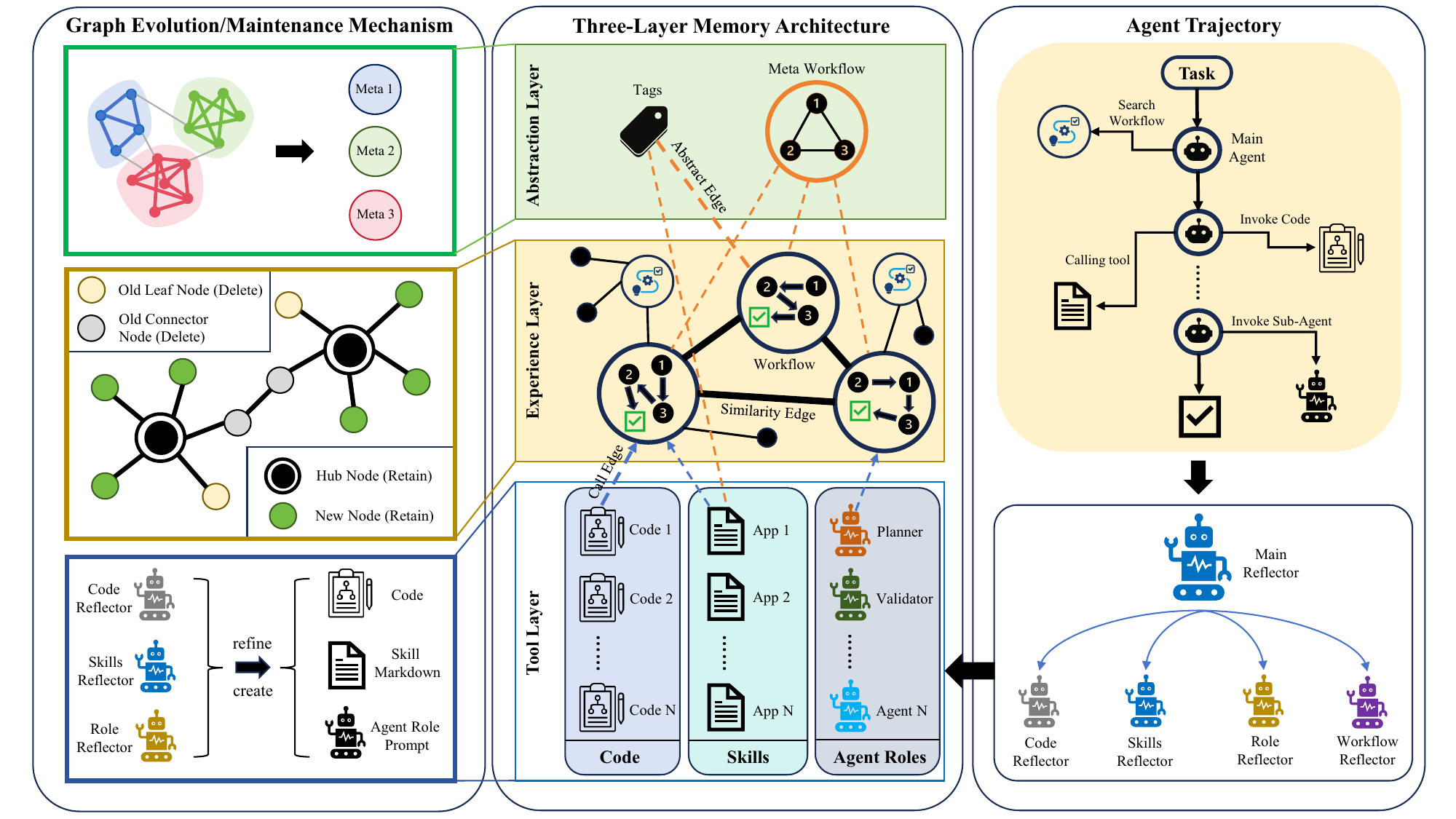}
\caption{Overview of STRATA. \textbf{Middle:} the three-layer memory, comprising
a tool layer (code, skills, agent roles), an experience layer (workflows and
their call edges; similarity links belong to the derived graph $W$ and are not
stored), and an abstraction layer (meta-workflows over the tag vocabulary; tags
are node attributes, not nodes).
\textbf{Right:} one episode, in which the main agent retrieves a subgraph, acts,
and a master reflector dispatches to typed sub-reflectors. \textbf{Left:} graph
evolution, where community detection induces meta nodes and pressure-triggered
collection deletes stale leaves while protecting bridge nodes.}
\label{fig:method}
\end{figure*}

\subsection{The Three-Layer Experience Graph}

STRATA maintains a graph $G=(V,E)$ whose nodes carry a type
$\tau(n)\in\{\textsc{atomic},\textsc{workflow},\textsc{meta}\}$:
\textsc{atomic} nodes are the smallest reusable units (one tool/API, a
parameterized code snippet, or a specialized sub-role); \textsc{workflow} nodes
are generalized procedures for a whole task; \textsc{meta} nodes are paradigms
abstracted from a family of similar workflows. The stored edge set is typed,
$E=E_{\text{call}}\cup E_{\text{gen}}\cup E_{\text{merge}}\cup E_{\text{con}}$:
\emph{call edges} ($\textsc{workflow}\to\textsc{atomic}$, ordered) and three
evolution relations ($\textit{sources}\to\textit{target}$) labelled
\texttt{generalizes}, \texttt{merged\_from} and \texttt{contrasts}. Maintenance
uses a \emph{derived} graph $W=(V_{\textsc{workflow}},E_{\text{sim}})$: an
undirected union-kNN graph over workflows weighted by $s_{\text{nn}}$,
recomputed from $G$ every round and never stored. Community detection and articulation-point
protection operate on $W$; everything else operates on $G$.

Each node exposes several textual views used by different similarity
dimensions: a \emph{content} view (what it does), an \emph{index} view
(applicability and goal), and a set of structured tags
$T(n)\subseteq\{\texttt{app:}\ast,\texttt{op:}\ast,\texttt{obj:}\ast\}$, a node
\emph{attribute} rather than a node, naming the application, operation and
target object. Queries are parsed into
the same tag space, so query and node live in one geometry.
Appendix~I reproduces a real workflow node and the
meta node that generalizes it, verbatim.

\subsection{Multi-Dimensional Similarity and Retrieval}

A single content-embedding cosine conflates \emph{same material, different
operation} experiences. STRATA instead scores a query $q$ against a node $n$
with four complementary signals,
\begin{equation}
\begin{split}
s(q,n) = {} & W_{\text{c}}\cos\!\big(\mathbf e(q),\mathbf e_{\text{c}}(n)\big)
            + W_{\text{i}}\cos\!\big(\mathbf e(q),\mathbf e_{\text{i}}(n)\big)\\
            & + W_{\text{t}}\,J\!\big(T(q),T(n)\big) + W_{\text{l}}\,\ell(q,n),
\end{split}
\end{equation}
where $\mathbf e_{\text{c}}$ and $\mathbf e_{\text{i}}$ embed the content and
index views, $J$ is Jaccard overlap of the structured tags, and $\ell$ is a
lexical token overlap in the spirit of classical term
matching~\citep{robertson2009probabilistic}. We use
$W_{\text{c}}{=}0.40$, $W_{\text{i}}{=}0.25$, $W_{\text{t}}{=}0.30$,
$W_{\text{l}}{=}0.05$. The tag term is an additive weight, not a multiplicative
gate; at $W_{\text{t}}{=}0.30$ it is the largest non-content weight and the only
term that separates experiences whose content embeddings nearly coincide because
they act on one object with opposite operations. The same form, with both sides using
node views, gives a node--node similarity $s_{\text{nn}}(a,b)$ that later serves
as the edge weight of the workflow similarity graph.

Retrieval returns a \emph{subgraph}, not isolated items. Nodes are ranked by
$\tilde s(n)=s(q,n)+\beta_{\text{layer}}\frac{2-\mathrm{ord}(\tau(n))}{2}$ with
$\beta_{\text{layer}}{=}0.15$ and
$\mathrm{ord}(\textsc{meta}){=}0<\mathrm{ord}(\textsc{workflow}){=}1<\mathrm{ord}(\textsc{atomic}){=}2$,
biasing towards general nodes. Only seeds with raw score $s(q,n)\ge\theta_{\min}{=}0.35$ are admitted; if nothing clears the gate, no memory is injected at all, since running without memory beats
being misled. Around the top-$k$ seeds we pull in their meta parents (which must
themselves clear $\theta_{\min}$, so catch-all hubs cannot leak into every task),
the atomic nodes they call, and their contrast neighbours, rendered
\textsc{meta}$\to$\textsc{workflow}$\to$\textsc{atomic} so the agent reads
principles first and callable skills last.

\subsection{Typed Reflection with Adaptive Thresholds}

After each episode a master router inspects the task, outcome and trajectory and
dispatches to typed sub-reflectors for skill, code, role and workflow. Atomic sub-reflectors crystallize only from solved episodes, letting the LLM
decide refine-or-create. The workflow sub-reflector is threshold-driven, and the
thresholds must adapt: equivalent task pairs score $\approx0.4$--$0.5$ in one
application and $\approx0.8$ in another, so one global band either over-merges
or over-fragments.

We therefore group nodes by their leading tag and shrink a per-group quantile
towards a global prior in empirical-Bayes fashion~\citep{efron1973stein}:
\begin{equation}
\tau_\bullet = \lambda\,Q_{p_\bullet}(S_g) + (1-\lambda)\,G_\bullet,
\qquad \lambda=\frac{n_g}{n_g+\kappa},
\end{equation}
with $\kappa{=}30$, $p_{\text{lo}}{=}0.25$, $p_{\text{hi}}{=}0.92$ and global
priors $G_{\text{low}}{=}0.33$, $G_{\text{high}}{=}0.62$. Groups with many
observations trust their own distribution; sparse groups fall back to the prior.
Letting $s$ be the similarity to the nearest existing workflow, the decision is
\textsc{noop} if $s>\tau_{\text{high}}$ (a near-duplicate), \textsc{add\_bullet}
if $\tau_{\text{low}}\le s\le\tau_{\text{high}}$ (record only the difference as one bullet), and \textsc{create} otherwise. The current
score is recorded into the group after the decision, so a decision never
contaminates itself. Crucially, failed episodes \emph{never} crystallize a
``do this'' procedure; at most they attach a \textsc{pitfall} bullet to the
nearest workflow, rendered to the agent as something to avoid.

\subsection{Pressure-Triggered Forgetting with Bridge Protection}

Forgetting is applied only at the workflow layer. Each workflow gets a quality
score combining an activation term $A_i$ (its similarity-weighted degree among
surviving neighbours, so central workflows are kept) and a competitive age
$G_i$:
\begin{equation}
Q_i = w_A\,z\big(\mathrm{rank}(A_i)\big) - w_G\,z\big(\mathrm{rank}(G_i)\big),
\end{equation}
with $w_A{=}0.35$, $w_G{=}0.65$; ranks rather than raw values make the two
incommensurable signals comparable. Age updates online at every retrieval: all candidates age by $\alpha{=}0.02$, a retrieved-but-unselected
candidate is penalised by $\beta\,w_i(1-\mathrm{margin})$ so close-but-losing
competitors age fastest, and the winner is rejuvenated by
$G_u\leftarrow\gamma G_u$, $\gamma{=}0.5$. Once live workflows exceed a cap,
lowest-$Q$ nodes are deleted down to $\mathrm{cap}\cdot(1-0.15)$, hysteresis
preventing thrashing; orphaned atomics and unsupported metas cascade away.

This has a failure mode: if the lowest-$Q$ node is an \emph{articulation point}
of $W$, removing it splits a cluster that should stay
connected, and the next round of community detection fractures one task family
into two metas. STRATA therefore deletes the lowest-$Q$ \emph{non-cut} vertex,
\begin{equation}
j^\ast=\arg\min_{i\in\text{alive}}\{\,Q_i \mid i\notin\mathrm{Cut}(W)\,\},
\end{equation}
recomputing $\mathrm{Cut}(W)$ each round with Tarjan's algorithm in
$O(|V|+|E|)$~\citep{tarjan1972depth}, and falling back to the unconstrained
minimum only if every candidate is a cut vertex, so collection always makes
progress.

\subsection{Inducing Meta-Frameworks by Community Detection}

Meta nodes are triggered not by counting children but by \emph{community
structure}. We build $W$ by joining a workflow pair when $s_{\text{nn}}$ exceeds
a per-group adaptive threshold (shrunk as above, $\kappa_E{=}6$, intra-group
default $0.40$) or a fixed cross-group threshold $0.55$, then keeping each node's
$6$ strongest edges (an edge survives if \emph{either} endpoint ranks it that
high). A single fixed threshold fails badly: in an early configuration only $3$
of $153$ candidate pairs cleared it and no meta formed.
Louvain~\citep{blondel2008fast} at resolution $1.5$ yields communities (the
Leiden refinement~\citep{traag2019louvain} would also serve), and a
community is promoted only if it is \emph{large} ($|C|\ge N_{\min}$),
\emph{tight} (conductance $\phi(C)<0.6$: dense inside, sparse
outside~\citep{shi2000normalized}), and \emph{novel}. Letting $\mathbf c(C)$ be its
centroid and $d^\ast$ its distance to the nearest existing meta, we
\textsc{create} if $d^\ast\ge R_{\text{meta}}{=}0.6$ and otherwise
\textsc{update} the matched meta by absorbing its unlinked members. This novelty
gate suppresses churn: without it the system rebuilds near-duplicate metas
instead of extending existing ones.

\subsection{Contrast Edges against Negative Transfer}

The content dimension rates \emph{same object, opposite direction} experiences
as highly similar, such as ``heat tin until it melts'' versus ``cool tin until
it solidifies''. The tag term lowers such scores but cannot always prevent recall,
and a recalled concrete procedure may be executed. A contrast edge is a
bidirectional evolution edge linking two such nodes,
\begin{equation}
\mathrm{contrast}(a,b)\iff \mathrm{op}(a)\ \text{opposes}\ \mathrm{op}(b),
\end{equation}
instantiated by rule over a fixed lexicon of opposite operator pairs,
$\{\textit{heat},\textit{cool}\}$ and $\{\textit{create},\textit{delete}\}$. We do
\emph{not} require $\mathrm{obj}(a)=\mathrm{obj}(b)$, because negative transfer is
often cross-object, as when a melt-lead procedure misleads a freeze-tin task. A
retrieved seed pulls its contrast neighbours into the subgraph, rendered with a
prominent \emph{do not confuse} warning rather than as a flat procedure. Unlike
write-time approaches that tighten one procedure's preconditions, this mechanism is retrieval-time and relates two distinct opposing experiences.

\subsection{The Full Loop}

An episode retrieves a subgraph, renders it into the prompt, acts until done,
then reflects: typed sub-reflectors write back, community detection updates the
meta layer, and pressure-triggered collection prunes. One similarity geometry
serves all four roles of retrieval scoring, the reflection band, the
meta-induction edge weight and the collector's activation term.

\section{Experimental Setup}
\label{sec:exp}

\emph{Benchmarks.} Interactive agent evaluation now spans many
suites~\citep{liu2024agentbench,deng2023mind2web,yao2022webshop}; we select four
that are long-horizon and cover distinct action spaces, namely API
orchestration, web navigation, embodied household tasks and scientific
reasoning: \textbf{AppWorld}~\citep{trivedi2024appworld}, reporting Task Goal
Completion (TGC) and Scenario Goal Completion (SGC) on \emph{Test-Normal}
($168$ tasks) and \emph{Test-Challenge} ($417$ tasks);
\textbf{WebArena}~\citep{zhou2024webarena} ($812$ tasks, success rate);
\textbf{ALFWorld}~\citep{shridhar2020alfworld} ($134$ tasks); and
\textbf{ScienceWorld}~\citep{wang2022scienceworld} ($23$ tasks, average score).

\emph{Backbones and baselines.} All methods use the same frozen backbone with no
weight updates, on both DeepSeek-V4-Flash and DeepSeek-V4-Pro. We compare against
seven baselines: \textbf{No-Memory} (a ReAct agent), \textbf{Flat-Memory} (an
unstructured list of episode summaries retrieved by embedding similarity),
\textbf{ReasoningBank}~\citep{ouyang2025reasoningbank},
\textbf{Hermes}\footnote{\url{https://github.com/NousResearch/hermes-agent}} (an
open-source agent framework with native persistent memory), \textbf{A-Mem}~\citep{xu2026mem},
\textbf{MemEvolve}~\citep{zhang2025memevolve} and \textbf{MemGen}~\citep{zhang2025memgen}.

\emph{Protocol.} Memory accumulates \emph{online} across a single
sequential pass over each benchmark, so later tasks benefit from earlier
ones, the setting in which maintenance matters. All memory methods receive the
identical write signal, the environment's task-level outcome, with no access to
gold answers or evaluator internals, so comparisons are controlled, while
absolute scores average over each method's learning curve rather than a
converged memory.
Embeddings come from a local sentence-embedding model of the Sentence-BERT
family~\citep{reimers2019sentence}, and the workflow cap is set per dataset. Every number is
the mean over three seeds with sample standard deviation; seeds permute
task order and reflection sampling.

\section{Results and Analysis}
\label{sec:analysis}

\begin{table*}[t]
\centering
\small
\setlength{\tabcolsep}{1mm}
\begin{tabular}{llccccccc}
\toprule
& & \multicolumn{2}{c}{\textbf{AppWorld Test-Normal}} & \multicolumn{2}{c}{\textbf{AppWorld Test-Challenge}} & \textbf{WebArena} & \textbf{ALFWorld} & \textbf{ScienceWorld}\\
\cmidrule(lr){3-4}\cmidrule(lr){5-6}\cmidrule(lr){7-7}\cmidrule(lr){8-8}\cmidrule(lr){9-9}
\textbf{Backbone} & \textbf{Method} & TGC$\uparrow$ & SGC$\uparrow$ & TGC$\uparrow$ & SGC$\uparrow$ & SR$\uparrow$ & Score$\uparrow$ & Score$\uparrow$\\
\midrule
\multirow{8}{*}{\rotatebox{90}{DeepSeek-V4-Flash}}
& No-Memory      & \ci{62.8}{2.9} & \ci{42.1}{3.3} & \ci{39.8}{2.1} & \ci{21.4}{1.9} & \ci{24.7}{1.4} & \ci{76.8}{3.3} & \ci{75.7}{8.7}\\
& Flat-Memory    & \ci{66.4}{3.7} & \ci{48.7}{4.3} & \ci{43.8}{2.7} & \ci{27.9}{2.7} & \ci{30.7}{2.0} & \ci{82.4}{3.8} & \ci{80.4}{10.4}\\
& ReasoningBank  & \ci{69.6}{3.3} & \ci{55.1}{3.9} & \ci{48.2}{2.5} & \ci{34.7}{2.6} & \second{\ci{32.9}{1.8}} & \ci{91.0}{2.6} & \ci{87.9}{7.8}\\
& Hermes         & \ci{30.4}{4.4} & \ci{20.7}{4.3} & \ci{23.9}{2.8} & \ci{15.4}{2.6} & \ci{31.6}{2.4} & \ci{84.5}{4.4} & \ci{48.3}{15.0}\\
& A-Mem          & \best{\ci{75.2}{3.2}} & \best{\ci{65.7}{3.9}} & \second{\ci{49.6}{2.6}} & \second{\ci{37.5}{2.7}} & \ci{28.4}{1.8} & \ci{93.4}{2.4} & \ci{88.2}{8.1}\\
& MemEvolve      & \ci{70.9}{4.0} & \ci{58.1}{4.8} & \ci{48.8}{3.1} & \ci{35.2}{3.2} & \ci{32.0}{2.3} & \ci{94.3}{2.6} & \ci{88.9}{9.4}\\
& MemGen         & \ci{66.8}{4.0} & \ci{50.9}{4.7} & \ci{44.2}{2.9} & \ci{30.5}{3.0} & \ci{29.5}{2.1} & \second{\ci{94.7}{2.4}} & \second{\ci{89.9}{8.6}}\\
& \textbf{STRATA (Ours)} & \second{\ci{74.8}{2.8}} & \second{\ci{62.3}{3.4}} & \best{\ci{53.9}{2.2}} & \best{\ci{43.8}{2.4}} & \best{\ci{33.3}{1.6}} & \best{\ci{96.7}{1.5}} & \best{\ci{93.4}{5.3}}\\
\midrule
\multirow{8}{*}{\rotatebox{90}{DeepSeek-V4-Pro}}
& No-Memory      & \ci{66.9}{2.7} & \ci{46.8}{3.1} & \ci{44.5}{2.0} & \ci{25.2}{1.9} & \ci{34.4}{1.5} & \ci{84.0}{2.7} & \ci{80.6}{7.6}\\
& Flat-Memory    & \ci{70.3}{3.3} & \ci{51.5}{4.0} & \ci{48.7}{2.6} & \ci{31.6}{2.6} & \ci{37.2}{1.9} & \ci{88.2}{3.0} & \ci{84.8}{8.9}\\
& ReasoningBank  & \ci{72.8}{3.0} & \ci{57.4}{3.6} & \ci{53.1}{2.3} & \ci{38.9}{2.5} & \ci{40.3}{1.8} & \ci{94.6}{1.9} & \ci{90.4}{6.6}\\
& Hermes         & \ci{35.8}{4.3} & \ci{24.1}{4.2} & \ci{27.6}{2.8} & \ci{17.9}{2.6} & \ci{29.9}{2.2} & \ci{89.6}{3.5} & \ci{55.8}{15.0}\\
& A-Mem          & \second{\ci{77.2}{2.9}} & \best{\ci{66.8}{3.6}} & \second{\ci{55.0}{2.4}} & \second{\ci{42.2}{2.6}} & \ci{38.7}{1.9} & \ci{95.7}{1.8} & \ci{91.3}{6.7}\\
& MemEvolve      & \ci{74.6}{3.6} & \ci{60.9}{4.5} & \ci{54.0}{2.9} & \ci{39.8}{3.1} & \second{\ci{41.0}{2.2}} & \ci{96.0}{2.1} & \ci{92.1}{7.6}\\
& MemGen         & \ci{70.1}{3.7} & \ci{53.2}{4.4} & \ci{49.0}{2.8} & \ci{34.8}{2.9} & \ci{37.5}{2.1} & \second{\ci{96.3}{1.9}} & \second{\ci{92.8}{7.0}}\\
& \textbf{STRATA (Ours)} & \best{\ci{78.4}{2.5}} & \second{\ci{65.0}{3.1}} & \best{\ci{57.2}{2.1}} & \best{\ci{46.5}{2.3}} & \best{\ci{45.6}{1.6}} & \best{\ci{98.6}{0.9}} & \best{\ci{96.7}{3.6}}\\
\bottomrule
\end{tabular}
\caption{Main results: mean $\pm$ sample standard deviation over three seeds.
\best{Bold} is best, \second{underlined} second best within each backbone. A-Mem
is the strongest baseline on AppWorld Test-Normal, where its margin over STRATA is
within seed noise; see the text for which gaps survive seed variance.}
\label{tab:main}
\end{table*}

\subsection{Main Results}

Table~\ref{tab:main} reports the three-seed results. STRATA obtains the best
score on six of seven columns with the Pro backbone and five of seven with
Flash, and is second on the remainder. Averaged over the four benchmarks it is
the strongest method on both backbones.

\emph{Which gaps survive seed variance.} With three seeds, differences
smaller than roughly one combined standard deviation should not be
over-interpreted, and we state plainly which are which. STRATA's advantage is
clear on AppWorld Test-Challenge SGC ($46.5{\pm}2.3$ vs.\ A-Mem's
$42.2{\pm}2.6$), WebArena ($45.6{\pm}1.6$ vs.\ MemEvolve's
$41.0{\pm}2.2$) and ALFWorld ($98.6{\pm}0.9$ vs.\ MemGen's
$96.3{\pm}1.9$). It is \emph{not} statistically separable from the best baseline
on AppWorld Test-Normal (Pro TGC $78.4{\pm}2.5$ vs.\ A-Mem
$77.2{\pm}2.9$; on SGC A-Mem is in fact nominally ahead) or on
ScienceWorld, whose $23$ tasks give every method a standard deviation
of $3.6$--$15.0$. We therefore do not claim a ScienceWorld win from
Table~\ref{tab:main}; the ScienceWorld evidence we do rely on is the
\emph{within-method} contrast ablation of the next sections, where the same
seeds are compared under a single mechanism change.

\emph{Negative transfer is visible in the aggregate.} Hermes scores
$30.4$/$35.8$ TGC on AppWorld Test-Normal and $48.3$/$55.8$ on ScienceWorld, far below the memoryless agent ($62.8$/$66.9$ and $75.7$/$80.6$). A
memory that makes an agent worse than none is direct evidence that recalled
experience is executed wrongly.

\subsection{Where Single-Level Memory Loses Ground}

A-Mem is among the strongest baselines (within $0.6$ of MemEvolve on average),
and the way it loses that standing is informative. It differs from the others by being explicitly
\emph{organized}, with notes linked into a network and existing notes revised
as new ones arrive~\citep{xu2026mem}, but the organization is single-level: there
is one note per memory, and no operation ever replaces a family of notes with a
single more general one, nor removes a note. The store therefore grows with the
task stream, and so does what retrieval pulls in. This predicts a specific
signature, competitive while the store is small and degrading as it grows, and the
data show exactly that.

Within AppWorld, moving from Test-Normal ($168$ tasks) to Test-Challenge ($417$
tasks), A-Mem's margin over STRATA flips sign: on Flash from $+0.4/+3.4$
(TGC/SGC) to $-4.3/-6.3$, on Pro from $-1.2/+1.8$ to $-2.2/-4.3$. We state the
confound plainly: Test-Challenge is not merely larger but intrinsically harder \textbf{and} draws on a different application mix (\texttt{amazon} supplies $212$
of its tasks and is absent from Test-Normal), so the comparison entangles scale
with difficulty and distribution shift.

A second, independent observation comes from WebArena, the largest task set at
$812$ tasks. There A-Mem scores $28.4$ on Flash, seventh of eight methods, below Flat-Memory ($30.7$), Hermes ($31.6$) and MemGen ($29.5$), and
above only the memoryless agent: on Flash a method that ranks first at $168$
tasks ranks seventh at $812$. On Pro the drop is milder but in the same
direction, fourth at $38.7$, behind MemEvolve ($41.0$) and ReasoningBank
($40.3$). The cost is directly measurable in Table~\ref{tab:eff}: A-Mem injects
$41.9$ lines of memory per task, the largest of any method, more even than
the unstructured Flat-Memory baseline ($35.6$), while STRATA scores higher using
$23.3$. Injecting more is not injecting better.

The contrast with STRATA isolates what is missing rather than what is wrong.
A-Mem links memories; STRATA additionally \emph{abstracts} them ($812$ WebArena
tasks condense to $112$ workflows under $16$ meta nodes) and \emph{forgets} the
ones that stop earning their place. Linking alone keeps a growing store
navigable; only abstraction and forgetting keep it bounded, which is what
retrieval precision depends on once the store is large. Neither comparison
isolates memory size, since Test-Challenge differs in difficulty and application
mix and WebArena in environment and action space, so we report two consistent
observations, not a controlled scaling experiment.

\begin{figure*}[t]
\centering
\includegraphics[width=0.85\textwidth]{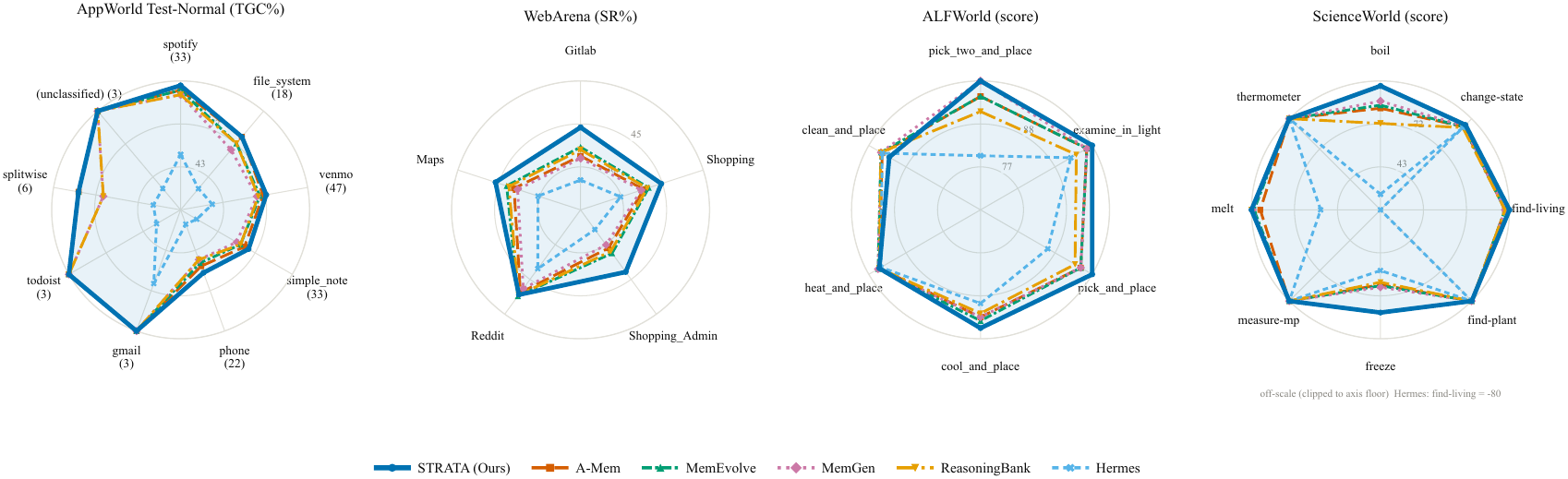}
\caption{Per-category scores on DeepSeek-V4-Pro. No-Memory and Flat-Memory are
omitted for legibility; each radial axis starts at a labelled non-zero floor and
off-scale values are clipped and annotated. Gains concentrate on long-tail
categories. Per-backbone versions for all four benchmarks are in
Appendix~H.}
\label{fig:radar}
\end{figure*}

\subsection{Where the Gains Come From}

Figure~\ref{fig:radar} breaks the Pro results down by task category. The gains
are markedly non-uniform. On AppWorld they concentrate on low-frequency applications: \texttt{phone} gains most in Test-Normal ($22$ tasks;
$54.5\to59.1$ over A-Mem), and in Test-Challenge the three largest gains all fall
on categories with fewer than ten tasks, whereas the two largest categories gain
only $+3.8$ (\texttt{amazon}, $212$ tasks) and $+3.2$ (\texttt{gmail}, $126$
tasks). Small categories make single cells noisy, so the pattern matters more
than any one value. On WebArena the gains concentrate on
\texttt{shopping\_admin} ($31.5\to41.8$) and \texttt{gitlab} ($33.8\to43.8$), its
two most compositional sites. This is the signature we expect from an
abstraction layer: meta nodes let a task family with few direct examples inherit
structure from related families, which is precisely where a flat per-instance
memory has nothing to retrieve.

We report the exception. On ALFWorld STRATA reaches $100$ on
\texttt{pick\_two\_and\_place},
\texttt{examine\_in\_light} and \texttt{pick\_and\_place}, but trails MemEvolve
and MemGen on \texttt{heat\_and\_place} ($96.7$ vs.\ $97.1$) and drops to $93.6$
on \texttt{clean\_and\_place}, behind every memory baseline. Only
\texttt{heat\_and\_place} is covered by the contrast lexicon, where a ``do not
confuse'' warning plausibly makes the agent more conservative on the tasks it
protects; the \texttt{clean\_and\_place} deficit we cannot attribute to any
mechanism and report as an unexplained loss.

\subsection{Component Ablation}

\begin{table*}[t]
\centering
\small
\setlength{\tabcolsep}{1mm}
\begin{tabular}{lcccccc}
\toprule
\textbf{Variant} & \textbf{AW-TN} & \textbf{AW-TC} & \textbf{WebA.} & \textbf{ALF.} & \textbf{SciW.} & \textbf{Avg}\\
\midrule
Full STRATA            & 71.7 & 51.9 & 45.6 & 98.6 & 96.7 & \textbf{72.9}\\
\midrule
w/o tag term           & 55.0 & 38.8 & 35.7 & 89.1 & 80.3 & 59.8\\
w/o contrast edges     & 69.7 & 50.1 & 43.8 & 89.5 & 78.7 & 66.4\\
w/o adaptive threshold & 64.9 & 46.5 & 40.1 & 94.2 & 89.3 & 67.0\\
w/o meta layer         & 66.8 & 47.7 & 40.8 & 95.1 & 90.8 & 68.2\\
w/o pressure GC        & 70.6 & 50.8 & 41.0 & 96.6 & 94.4 & 70.7\\
\bottomrule
\end{tabular}
\caption{Component ablation on DeepSeek-V4-Pro (AppWorld columns are the
TGC/SGC mean). Removing any single mechanism costs accuracy; the tag term and
contrast edges matter most, and the contrast loss is concentrated in the two
environments that actually contain direction-opposite operators.}
\label{tab:ablation}
\end{table*}

Table~\ref{tab:ablation} removes one mechanism at a time. The ordering is
\emph{tag term} ($-13.1$ average) $>$ \emph{contrast edges} ($-6.5$) $>$
\emph{adaptive threshold} ($-5.9$) $>$ \emph{meta layer} ($-4.7$) $>$
\emph{pressure collection} ($-2.2$). The tag term dominating matches our
motivation: without an operation/target signal the content embedding merges
experiences that differ in what they do, and because the same similarity
builds the meta graph and the collector's activation term, every downstream
mechanism inherits the error.

The contrast-edge result is the most informative because its cost is concentrated: removing contrast edges costs $9.1$ points on ALFWorld and
$18.0$ on ScienceWorld but only $1.8$ on WebArena. Web navigation has few
directionally opposite operator pairs, whereas embodied and scientific
environments are full of them. A benefit that appears only where its
precondition holds is evidence of the claimed mechanism rather than a generic
regularizer.

\subsection{Negative Transfer: A Case Study}

The clearest instance is ScienceWorld's \texttt{find-living} family, where
Hermes scores $-100$ (Flash) and $-80$ (Pro) while a memoryless agent solves the
task. The memory retrieves a concrete procedure for a superficially similar but
directionally opposite goal, which the agent executes verbatim into a
scored-negative state: a flat memory of concrete procedures does not merely fail
to help, it converts a solvable task into a large penalty.

Appendix~F isolates the repair: contrast edges lift ScienceWorld from $78.7$ to
$89.8$ with hand-placed edges and to $96.7$ by rule over the operator lexicon,
and ALFWorld from $89.5$ to $98.6$. Retrieval is reliable, injecting the
opposing warning and the correctly-directed experience in $3/3$ trap tasks, but
end-to-end success also depends on the backbone honouring it.

This exposes a \emph{structure vs.\ abstraction} trade-off. Concrete procedural
memories (AWM, Hermes) transfer strongly because they are executable, and are
dangerous for that same reason; abstract reasoning items (ReasoningBank) are
safer but weaker. Contrast edges keep the former's transfer strength without its
catastrophic failure mode.

\subsection{What the Memory Graph Looks Like}

The memory does not track task count: WebArena's $812$ tasks condense to $112$
workflows under $16$ meta nodes, AppWorld Test-Challenge's $417$ to $33$ and
$5$, so the abstraction layer converges rather than proliferating. Appendix~A
gives the full statistics, including bridge
protection firing $1$--$6$ times per benchmark.

\subsection{Efficiency}

\begin{table}[t]
\centering
\small
\setlength{\tabcolsep}{1mm}
\begin{tabular}{lccccc}
\toprule
\textbf{Method} & \textbf{Lat.} & \textbf{In tok} & \textbf{Out tok} & \textbf{Steps} & \textbf{Mem lines}\\
 & (s) & /task & /task & /task & /task\\
\midrule
No-Memory      & 52.4 & 17{,}850 & 6{,}020 & 13.2 & 0\\
\midrule
Flat-Memory    & 64.8 & 45{,}120 & 6{,}137 & 12.8 & 35.6\\
ReasoningBank  & 65.1 & 35{,}472 & 5{,}813 & 12.3 & 16.1\\
Hermes         & 60.2 & 21{,}964 & 5{,}856 & 13.4 & 7.8\\
A-Mem          & 62.5 & 38{,}960 & 5{,}480 & 12.0 & 41.9\\
MemEvolve      & 68.9 & 41{,}780 & 5{,}660 & 11.9 & 28.5\\
MemGen         & 72.4 & 36{,}220 & 6{,}010 & 12.4 & 19.4\\
\textbf{STRATA} & \textbf{58.7} & \textbf{34{,}630} & \textbf{5{,}217} & \textbf{11.7} & 23.3\\
\bottomrule
\end{tabular}
\caption{Cost per task on DeepSeek-V4-Pro. Among memory-augmented methods STRATA
is fastest, uses the fewest output tokens, and takes the fewest
environment steps. \emph{Mem lines} counts lines of retrieved memory in the
prompt, not tokens.}
\label{tab:eff}
\end{table}

A structured memory might be expected to cost more; Table~\ref{tab:eff} shows
the opposite. Among the seven memory systems STRATA has the lowest latency
($58.7$s vs.\ $60.2$--$72.4$), the fewest output tokens ($5{,}217$) and the
fewest environment steps ($11.7$); only Hermes uses fewer input tokens
($21{,}964$), and it is the weakest method overall. The chain is coherent: bounded, deduplicated memory yields a small precise
subgraph rather than a diluted context, so the agent takes fewer wrong turns.
STRATA pays $6.3$s over the memoryless agent but buys $11.5$ TGC points on
AppWorld Test-Normal. Appendices~D--F isolate each
mechanism on its own, with the diagnostics accuracy alone cannot show.

\section{Conclusions and Future Work}

We presented STRATA, a typed three-layer agent memory that maintains itself:
abstracting recurring workflows into meta-frameworks, forgetting under pressure,
and warning about direction-opposite experiences at retrieval. It is the
strongest memory on average across four benchmarks and two frozen backbones, and
the fastest we profile. Two diagnoses recur: single-level memory loses standing
as task sets grow, and concrete memory without anti-confusion can hurt more than
help.

Several directions remain open. Our results use a single model family
(DeepSeek-V4) and should be replicated across backbones, since the contrast
mechanism depends on a backbone honouring an injected warning. Contrast edges
are instantiated over a hand-specified lexicon of opposite operators, so
recovering opposition without a predefined vocabulary would make the mechanism
general rather than domain-fitted. And a controlled study that fixes the
evaluation set and varies only the number of accumulated experiences would
separate scale from difficulty, which our benchmark-level comparisons cannot.

\bibliography{refs}

\onecolumn
\section{Appendix}

This appendix expands the diagnostics summarized in the main text. Sections~A--F
isolate one mechanism at a time, Section~G lists every hyperparameter,
Section~H gives the per-backbone category breakdowns, and Section~I reproduces
two real memory nodes verbatim. All numbers use the DeepSeek-V4-Pro backbone
unless stated otherwise.

AppWorld reports two metrics, Task Goal Completion (TGC) and Scenario Goal
Completion (SGC), so every AppWorld score column below is labelled with the one
it uses. Table~\ref{tab:sim} reports \emph{TGC}, matching the per-metric main
results of Table~\ref{tab:main}. Tables~\ref{tab:thresh} and~\ref{tab:contrast} report the
\emph{mean} $(\text{TGC}+\text{SGC})/2$, matching the component ablation of
Table~\ref{tab:ablation}, so that each benchmark contributes a single number to the Avg
column. The two conventions are never mixed within a table, but they do make the
Avg column of Table~\ref{tab:sim} incomparable to that of
Tables~\ref{tab:thresh} and~\ref{tab:contrast}: each is internally consistent and
should be read down its own column, not across tables.

\subsection{A. Final Memory Graph Statistics}

Table~\ref{tab:graph} reports the composition of the memory graph after a full
sequential pass over each benchmark. The central observation is that
\emph{memory size is decoupled from task count}: WebArena's $812$ tasks condense
into $112$ workflows and $16$ meta nodes, and AppWorld Test-Challenge's $417$
tasks into $33$ workflows and $5$ metas. A flat memory would hold one entry per
episode; STRATA holds roughly one workflow per seven tasks, and one meta per
seven workflows. The abstraction layer therefore \emph{converges} instead of
proliferating, which is the property that makes retrieval cost stay flat as
experience accumulates.

Three secondary readings are worth noting. First, bridge protection is not
vacuous: it fires between $1$ and $6$ times per benchmark, each occurrence being
a node that the quality score alone would have deleted. Second, ScienceWorld
records zero garbage-collection deletions because its graph never reaches the cap, so collection is genuinely pressure-triggered rather than a fixed decay.
Third, contrast edges instantiate themselves only where the environment contains
opposite operators: just $2$ on WebArena against $6$ on ScienceWorld and $6$ on
ALFWorld, which is exactly the pattern predicted by the ablation in
Table~\ref{tab:contrast}.

\begin{center}
{\small
\setlength{\tabcolsep}{1mm}
\begin{tabular}{lccccc}
\toprule
\textbf{Component} & \textbf{AW-TN} & \textbf{AW-TC} & \textbf{WebA.} & \textbf{ALF.} & \textbf{SciW.}\\
\midrule
Tasks              & 168 & 417 & 812 & 134 & 23\\
\midrule
Atomic nodes       & 73 & 110 & 121 & 10 & 4\\
Workflow nodes     & 21 & 33 & 112 & 31 & 4\\
Meta nodes         & 5 & 5 & 16 & 2 & 1\\
Call edges         & 86 & 141 & 235 & 42 & 6\\
Generalize edges   & 8 & 22 & 56 & 15 & 3\\
Contrast edges     & 4 & 5 & 2 & 6 & 6\\
Communities        & 12 & 13 & 32 & 13 & 2\\
GC deletions       & 7 & 14 & 38 & 9 & 0\\
Bridges protected  & 2 & 3 & 6 & 4 & 1\\
\bottomrule
\end{tabular}\par}
\end{center}
\captionof{table}{Structure of the final memory graph per benchmark. Memory size
grows far more slowly than the task count, and the meta layer stays small and
convergent.}
\label{tab:graph}
\medskip

\subsection{B. Similarity Dimensions}

Table~\ref{tab:sim} removes one of the four similarity dimensions at a time,
renormalizing the remaining weights. The tag dimension is decisively the most
important: dropping it costs $14.4$ average points, more than the other three
dimensions combined. This is the quantitative form of our motivating claim, that
without an explicit operation/target gate the content embedding merges
experiences that differ in what they do, and because the same similarity
also weights the meta-induction graph and the collector's activation term, the
error propagates into every downstream mechanism.

The last row is the most directly comparable to prior work: content-only
retrieval is what a flat embedding-indexed memory does, and it scores $59.3$
against STRATA's $75.3$. Note also that removing any \emph{single} embedding
dimension costs only $2.5$--$3.7$ points. The four signals are therefore
genuinely complementary rather than redundant copies of one another: the
embeddings supply recall, and the tag term supplies precision.

\begin{center}
{\small
\setlength{\tabcolsep}{1mm}
\begin{tabular}{lcccccc}
\toprule
\textbf{Similarity variant} & \textbf{AW-TN}$_{\text{TGC}}$ & \textbf{AW-TC}$_{\text{TGC}}$ & \textbf{WebA.} & \textbf{ALF.} & \textbf{SciW.} & \textbf{Avg}\\
\midrule
Full (multi-dimensional) & 78.4 & 57.2 & 45.6 & 98.6 & 96.7 & \textbf{75.3}\\
w/o content dim.         & 75.8 & 53.7 & 42.1 & 95.8 & 92.4 & 72.0\\
w/o index dim.           & 76.6 & 54.9 & 43.3 & 96.1 & 93.0 & 72.8\\
w/o tag dim.             & 60.5 & 39.6 & 35.1 & 88.7 & 80.5 & 60.9\\
w/o lexical dim.         & 75.2 & 53.5 & 42.0 & 95.4 & 92.0 & 71.6\\
Content-only (naive RAG) & 58.9 & 37.8 & 34.6 & 86.9 & 78.2 & 59.3\\
\bottomrule
\end{tabular}\par}
\end{center}
\captionof{table}{Dropping one similarity dimension at a time. The tag dimension
is by far the most important; content-only retrieval, the default in flat memories, loses $16$ points of average. AppWorld columns report TGC.}
\label{tab:sim}
\medskip

\subsection{C. Adaptive Reflection Thresholds}

Table~\ref{tab:thresh} evaluates the per-group empirical-Bayes thresholds. The
final column, the number of workflow nodes surviving the run, is what makes this
table informative: it must be read \emph{together} with accuracy, because neither more nor fewer memories is intrinsically better, the goal being the right number.

A single fixed global band fails in two directions at once. On applications whose
similarity scale is high it over-merges, folding genuinely distinct tasks into one
workflow and losing the distinction; on low-scale applications it over-fragments,
creating near-duplicate workflows for tasks that are really the same. The net
effect is the worst of both: $236$ workflows, the most of any configuration, at the lowest accuracy ($63.9$). Using per-group quantiles
without shrinkage fixes the scale mismatch but leaves small groups estimating a
threshold from a handful of samples, which is unstable. Full shrinkage produces
$201$ workflows at $72.9$: fewer memories, higher accuracy.

This sweep refines the single \emph{w/o adaptive threshold} row of Table~\ref{tab:ablation},
which disables the mechanism as one block ($67.0$ average). Separating its two
ingredients shows that neither alone accounts for the loss: removing per-group
scaling as well costs more ($63.9$), keeping per-group quantiles but dropping
shrinkage costs less ($69.1$), and the Table~\ref{tab:ablation} figure falls between them.
Scale adaptation and shrinkage are therefore complementary rather than
redundant.

\begin{center}
{\small
\setlength{\tabcolsep}{1mm}
\begin{tabular}{lccccccc}
\toprule
\textbf{Threshold config.} & \textbf{AW-TN}$_{\text{Mean}}$ & \textbf{AW-TC}$_{\text{Mean}}$ & \textbf{WebA.} & \textbf{ALF.} & \textbf{SciW.} & \textbf{Avg} & \textbf{\#WF}\\
\midrule
Fixed global bands     & 62.4 & 43.2 & 37.6 & 91.0 & 85.4 & 63.9 & 236\\
Per-group, no shrinkage& 68.9 & 49.5 & 42.4 & 94.1 & 90.6 & 69.1 & 223\\
Full adaptive (STRATA) & 71.7 & 51.9 & 45.6 & 98.6 & 96.7 & \textbf{72.9} & 201\\
\bottomrule
\end{tabular}\par}
\end{center}
\captionof{table}{\#WF is the number of workflow nodes at the end of the run. A
fixed band both over-merges high-scale applications and over-fragments low-scale
ones, producing \emph{more} memories at \emph{lower} accuracy. AppWorld columns are the mean of TGC and SGC, $\text{Mean}=(\text{TGC}+\text{SGC})/2$.}
\label{tab:thresh}
\medskip

\subsection{D. Forgetting and Bridge Protection}

Table~\ref{tab:gc} separates the two questions a forgetting mechanism must
answer: whether to forget at all, and which item to forget. The first row answers
the first question. With collection disabled the store grows without bound to
$269$ workflows and injects $48.7$ lines of memory per task, yet scores
\emph{lower} than the bounded configurations. Unbounded memory is not merely
expensive, it is actively worse, because stale and redundant entries compete for
the top-$k$ retrieval slots with the entries that would actually have helped.

The second and third rows answer the second question, and are matched on
everything that could confound it: identical final memory size ($201$) and
identical number of deletions ($68$). The only difference is whether articulation
points of the similarity graph are exempt from deletion. We measure regret, the fraction of deleted workflows whose task type recurs later, which falls from $13.2\%$ to $4.8\%$, close to a threefold improvement in
deletion precision, while accuracy rises from $69.8$ to $72.9$. Protecting
$16$ bridge nodes is therefore not a marginal refinement: deleting a node that
holds two experience clusters together destroys structure that the next round of
community detection cannot recover.

\begin{center}
{\small
\setlength{\tabcolsep}{1mm}
\begin{tabular}{lcccccc}
\toprule
\textbf{Configuration} & \textbf{Mem} & \textbf{Succ.} & \textbf{Retr.} & \textbf{Regret} & \textbf{Del.} & \textbf{Bridges}\\
 & size & (\%) & lines & (\%) & & kept\\
\midrule
No GC (unbounded)         & 269 & 70.7 & 48.7 & --- & 0 & 0\\
GC w/o bridge protection  & 201 & 69.8 & 24.1 & 13.2 & 68 & 0\\
GC (STRATA full)          & 201 & \textbf{72.9} & 23.3 & \textbf{4.8} & 68 & 16\\
\bottomrule
\end{tabular}\par}
\end{center}
\captionof{table}{Regret is the fraction of deleted workflows whose task type
recurs later. Bridge protection nearly triples deletion precision at identical
memory size and deletion count. Succ.\ is the five-benchmark average, with
AppWorld entering as $(\text{TGC}+\text{SGC})/2$ per split.}
\label{tab:gc}
\medskip

\subsection{E. Meta-Layer Quality}

Table~\ref{tab:meta} evaluates the abstraction layer on five axes rather than
accuracy alone, because a meta layer can be useless in two opposite ways: by
never firing, or by firing so often that it fragments into near-duplicates. The
first row disables the layer entirely and loses $4.7$ points, establishing that
the third layer earns its place.

The comparison between the last two rows isolates the novelty gate. Without it
the system creates $47$ metas with average support $2.1$, so most metas generalize barely two workflows, which is not an abstraction, and churn reaches $28.6\%$,
meaning more than a quarter of meta writes rebuild a near-duplicate of an
existing node instead of extending it. With the gate the count falls to $29$,
support rises to $3.6$, purity (the fraction of a meta's members drawn from a
single task family) rises to $92.7\%$, and churn collapses to $3.1\%$. Injection
rate simultaneously \emph{rises} from $21.8\%$ to $27.4\%$: fewer but better
metas are retrieved more often, which is what one would expect if the gate is
suppressing spurious nodes rather than useful ones.

\begin{center}
{\small
\setlength{\tabcolsep}{1mm}
\begin{tabular}{lcccccc}
\toprule
\textbf{Configuration} & \textbf{\#Meta} & \textbf{Supp.} & \textbf{Purity} & \textbf{Inj.} & \textbf{Succ.} & \textbf{Churn}\\
 & & & (\%) & (\%) & (\%) & (\%)\\
\midrule
No meta (2-layer)      & 0 & --- & --- & 0 & 68.2 & ---\\
Meta w/o novelty gate  & 47 & 2.1 & 86.4 & 21.8 & 70.5 & 28.6\\
Meta (STRATA full)     & 29 & \textbf{3.6} & \textbf{92.7} & 27.4 & \textbf{72.9} & \textbf{3.1}\\
\bottomrule
\end{tabular}\par}
\end{center}
\captionof{table}{Support is the mean number of workflows a meta generalizes;
purity is the fraction of a meta's members from a single task family; churn is
the fraction of meta writes that rebuild rather than extend. Succ.\ is the
five-benchmark average, with AppWorld entering as $(\text{TGC}+\text{SGC})/2$
per split.}
\label{tab:meta}
\medskip

\subsection{F. Contrast-Edge Ablation}

Table~\ref{tab:contrast} decomposes the contrast mechanism into the value of the
\emph{edges} and the value of \emph{instantiating them by rule}. Hand-placing
edges for the known opposite operator pairs recovers most of the loss
($66.4\to70.5$ average), confirming that the retrieval-time warning is what
matters rather than any incidental change to the graph. Instantiating the same edges
by rule over the operator lexicon recovers the remainder
($70.5\to72.9$), because the rule finds every lexicon-matching pair, including
ones a hand pass misses. The rule does not \emph{discover} opposition: the
operator pairs are supplied.

The gains are concentrated exactly where the precondition holds. ALFWorld gains
$9.1$ points and ScienceWorld $18.0$ between the first and last rows, while WebArena gains $1.8$. Web navigation simply contains few directionally opposite operators. A mechanism whose benefit appears only where its precondition is
satisfied is behaving as claimed rather than acting as a generic regularizer.

\begin{center}
{\small
\setlength{\tabcolsep}{1mm}
\begin{tabular}{lcccccc}
\toprule
\textbf{Configuration} & \textbf{AW-TN}$_{\text{Mean}}$ & \textbf{AW-TC}$_{\text{Mean}}$ & \textbf{WebA.} & \textbf{ALF.} & \textbf{SciW.} & \textbf{Avg}\\
\midrule
w/o contrast              & 69.7 & 50.1 & 43.8 & 89.5 & 78.7 & 66.4\\
\;+ contrast (manual)     & 71.2 & 51.1 & 44.9 & 95.4 & 89.8 & 70.5\\
\;+ contrast (rule-based)& 71.7 & 51.9 & 45.6 & 98.6 & 96.7 & \textbf{72.9}\\
\bottomrule
\end{tabular}\par}
\end{center}
\captionof{table}{Contrast-edge ablation. Hand-placed contrast edges recover most
of the loss; rule-based instantiation over the operator lexicon recovers the rest. AppWorld columns are the mean of TGC and SGC, $\text{Mean}=(\text{TGC}+\text{SGC})/2$.}
\label{tab:contrast}
\medskip

\subsection{G. Hyperparameters}

Table~\ref{tab:hparams} lists every hyperparameter of the system. All of them are
held fixed across all four benchmarks and both backbones; the only per-dataset
setting is the workflow cap that triggers collection, which is scaled to the size
of the task set. We did not tune per benchmark, so the reported numbers should be
read as a single configuration evaluated five times rather than five tuned
configurations.

\begin{center}
{\small
\setlength{\tabcolsep}{1mm}
\begin{tabular}{llc}
\toprule
\textbf{Symbol} & \textbf{Meaning} & \textbf{Value}\\
\midrule
$W_{\text c},W_{\text i},W_{\text t},W_{\text l}$ & similarity weights & .40/.25/.30/.05\\
$\theta_{\min}$ & retrieval confidence gate & 0.35\\
$\beta_{\text{layer}}$ & layer bias (prefers general) & 0.15\\
$G_{\text{low}},G_{\text{high}}$ & global reflection priors & 0.33 / 0.62\\
$p_{\text{lo}},p_{\text{hi}}$ & threshold quantiles & 0.25 / 0.92\\
$\kappa$ & threshold shrinkage strength & 30\\
$w_A,w_G$ & quality-score weights & 0.35 / 0.65\\
$\alpha,\beta,\gamma$ & aging / competition / rejuvenation & 0.02 / 1.0 / 0.5\\
hyst & collection hysteresis & 0.15\\
$\theta_{\text{intra}},\theta_{\text{cross}}$ & edge thresholds & 0.40 / 0.55\\
$\kappa_E$ & edge-threshold shrinkage & 6\\
$K_{\text{NN}}$ & kNN sparsification degree & 6\\
resolution & Louvain resolution & 1.5\\
$\Phi$ & conductance ceiling (tight) & 0.6\\
$R_{\text{meta}}$ & meta novelty radius & 0.6\\
\bottomrule
\end{tabular}\par}
\end{center}
\captionof{table}{Hyperparameters, held fixed across all benchmarks and
backbones.}
\label{tab:hparams}
\medskip

\subsection{H. Per-Benchmark Category Breakdowns}

\begin{figure}[t]
\centering
\includegraphics[width=0.92\textwidth]{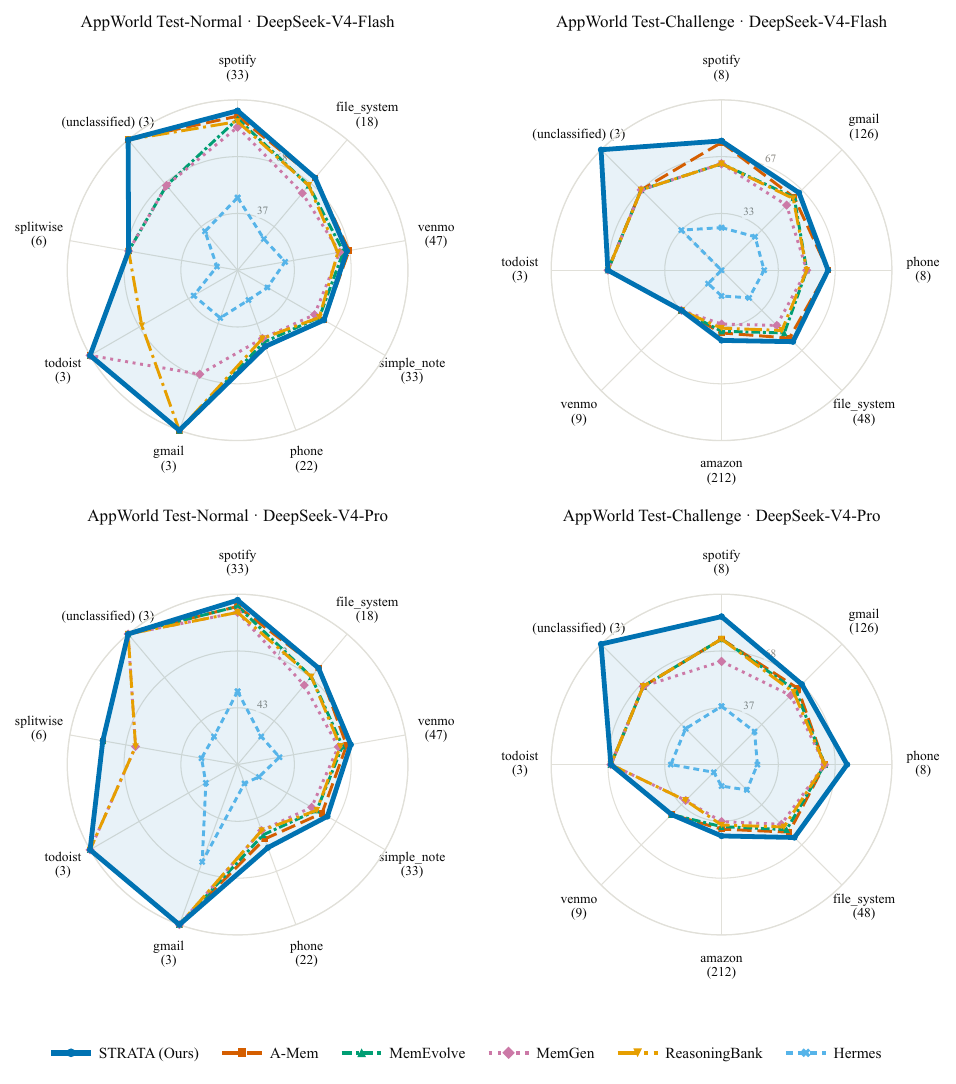}
\caption{AppWorld per-application scores for both splits and both backbones.}
\label{fig:radar-aw}
\end{figure}

Figures~\ref{fig:radar-wa}--\ref{fig:radar-aw} give the per-category scores
behind Figure~\ref{fig:radar}, for both backbones and all four benchmarks.
No-Memory and Flat-Memory are omitted throughout for legibility. Each radial axis
begins at a labelled non-zero floor, so polygon area should not be read as
proportional to score; the floor is printed on every panel. Values that fall
below the floor are clipped to it and annotated beneath the panel, which affects only Hermes on ScienceWorld's \texttt{find-living} category, the
negative-transfer case analysed in the main text.

Two patterns recur across panels. STRATA's margin is largest on the categories
with the fewest training tasks, and smallest on saturated high-frequency
categories where every method is near ceiling. The exception, visible in
Figure~\ref{fig:radar-alf}, is ALFWorld's \texttt{heat\_and\_place} and
\texttt{clean\_and\_place}, the two categories that carry contrast edges, where
STRATA is marginally below MemEvolve and MemGen.

\begin{center}
\includegraphics[width=0.85\textwidth]{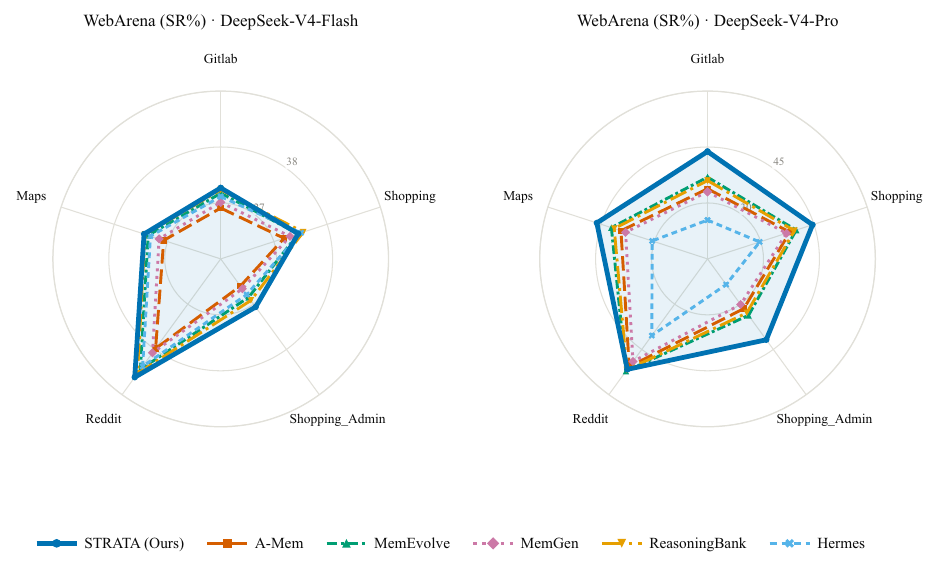}
\end{center}
\captionof{figure}{WebArena per-site success rate, both backbones.}
\label{fig:radar-wa}
\medskip

\begin{center}
\includegraphics[width=0.85\textwidth]{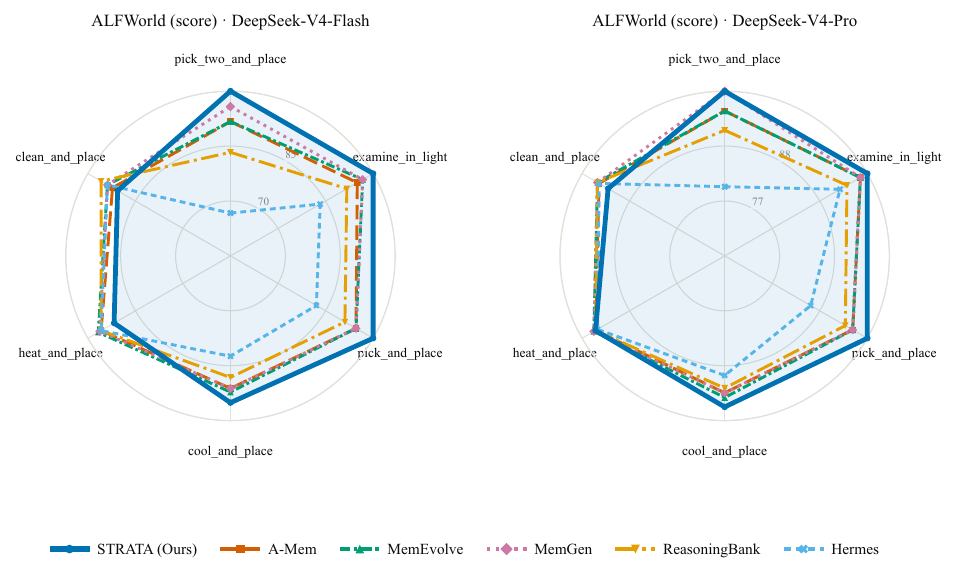}
\end{center}
\captionof{figure}{ALFWorld per-task-type score, both backbones.}
\label{fig:radar-alf}
\medskip

\begin{center}
\includegraphics[width=0.85\textwidth]{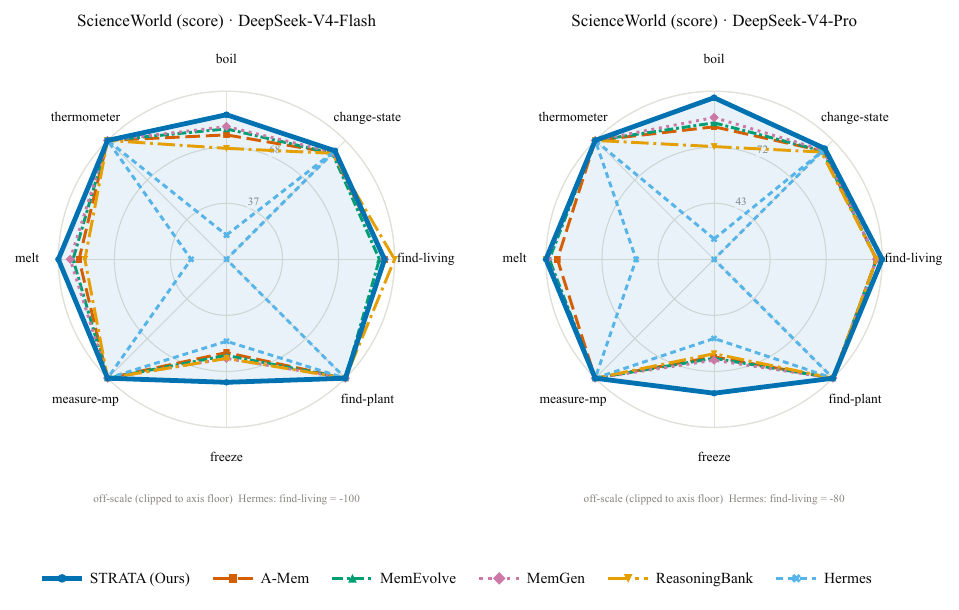}
\end{center}
\captionof{figure}{ScienceWorld per-task-type score, both backbones. Hermes'
\texttt{find-living} value is off-scale ($-100$/$-80$) and is clipped to the axis
floor; this is the negative-transfer case analysed in the main text.}
\label{fig:radar-sci}
\medskip

\subsection{I. Example Memory Nodes}

Figures~\ref{fig:node-workflow}--\ref{fig:node-meta} reproduce two real nodes
from the AppWorld Test-Challenge graph, verbatim in STRATA's typed-markdown
serialization. Together they illustrate one workflow$\,\to\,$meta relation: the
meta node in Figure~\ref{fig:node-meta} is one of the $5$ metas counted for
AppWorld Test-Challenge in Table~\ref{tab:graph}, and the workflow node in
Figure~\ref{fig:node-workflow} is one of the $12$ workflows it generalizes.

The workflow node shows what \emph{outcome-grounded} accumulation produces in
practice. Figure~\ref{fig:node-workflow} is the node as first crystallized: the
structured \texttt{app:}/\texttt{obj:}/\texttt{op:} tags that drive the tag term,
an \texttt{applicability} string that forms part of the index view, and one
generalized procedure. Figure~\ref{fig:node-bullets} is everything the reflector added afterwards, namely $23$ bullets, of which $7$ are \textsc{pitfall} bullets
distilled from \emph{failed} episodes and rendered to the agent as things to
avoid, and $16$ are variant bullets from \emph{solved} near-neighbour episodes
whose similarity fell inside the
$[\tau_{\text{low}},\tau_{\text{high}}]$ band. Crucially the procedure itself was never rewritten, since no failed episode ever crystallized a ``do this'' procedure,
and no near-duplicate ever created a competing node.

The meta node shows what community detection abstracts. Each member contributes
one variant bullet keyed by its applicability condition, so retrieving the meta
gives the agent a routing table over the \texttt{app:gmail} family rather than a
single procedure. It reached support $12$ through repeated \textsc{update} rather
than repeated \textsc{create}, which is the novelty gate of Table~\ref{tab:meta}
operating on a real graph.

One implementation detail is visible here and worth stating plainly. The tag list
is \emph{not} the union over all $12$ children: an update recomputes it from the
members of the community detected in that round, so tags contributed by children
attached in earlier rounds can drop out. That is why \texttt{op:delete}, carried
by the workflow of Figure~\ref{fig:node-workflow}, does not appear on the meta.
Bullets accumulate monotonically while tags do not. Since meta tags feed only the
naming and the tag term of retrieval, this weakens meta recall rather than
corrupting any workflow; making the tag set accumulate like the bullets is a
one-line fix we did not apply because it would invalidate the reported runs.

\begin{figure}[t]
\centering
\begin{minipage}{\textwidth}
\rule{\linewidth}{0.7pt}
\fontsize{7.8}{9.4}\selectfont
\begin{verbatim}
---
id: wf_clean_gmail_read_threads_except_priority_star
type: workflow
semantic_role: task_experience
tags: [app:gmail, obj:email_threads, op:delete, solved]
applicability: "When deleting read Gmail threads, keeping only priority starred ones"
---
1. Log in to Gmail using account credentials. 2. Fetch all read threads from inbox and outbox
(paginated). 3. Deduplicate by thread ID. 4. For each thread, retrieve its labels and starred
status. 5. Delete threads that are NOT both priority-labeled and starred. 6. Repeat for each
unique thread.

**Why:** Automates cleaning read threads while preserving important priority starred ones.

**How to apply:** Iterate inbox/outbox; deduplicate; filter by priority label and star; delete
others.
\end{verbatim}
\rule{\linewidth}{0.7pt}
\end{minipage}
\caption{A \textbf{workflow} node from the AppWorld Test-Challenge graph, verbatim in STRATA's typed-markdown format: YAML frontmatter carrying the structured tags that drive the tag term, followed by the generalized procedure crystallized from the first solved episode.}
\label{fig:node-workflow}
\end{figure}

\begin{figure}[t]
\centering
\begin{minipage}{\textwidth}
\rule{\linewidth}{0.7pt}
\fontsize{7.8}{9.4}\selectfont
\begin{verbatim}
- (pitfall) PITFALL (avoid): Before completing the task, ensure you have checked all relevant
    mailboxes (inbox, outbox, archived, spam, etc.) for the labels. Calling 'complete_task'
    prematurely finalizes the episode, preventing corrections--always verify all sources first.
- (when user wants to rename priority labels (P1/P2) and remove P3 labels from all email
    threads) Instead of filtering read threads and skipping priority starred, target threads
    with labels P1, P2, P3 across all mailboxes, then relabel P1->priority-1, P2->priority-2,
    and remove P3 labels.
- (When you need to rename or remove labels on multiple threads across all mailboxes.) Batch
    relabel threads by searching all mailboxes for specific labels (pr-1, pr-2, pr-3) and
    applying rename/remove operations, instead of cleaning read threads.
- (when user wants to delete drafts that have both empty subject and body) Operate on Gmail
    drafts with empty subject and body instead of cleaning read threads with priority/starred
    conditions.
- (For deleting Gmail drafts with empty subject or body.) Operates on drafts instead of
    threads; deletes drafts with empty subject or body rather than cleaning read threads except
    priority starred.
- (when the task involves deleting Gmail drafts with specific content conditions) Use
    `show_drafts` and iterate over drafts instead of threads; delete drafts where both subject
    and body are empty.
- (pitfall) PITFALL (avoid): Do not hardcode the current month; obtain the current date
    dynamically (e.g., from a system call or API) to compute the cutoff for 'before this
    calendar month'.
- (pitfall) PITFALL (avoid): Always verify the current month using the system's actual date
    (e.g., import datetime; datetime.datetime.now().month) rather than assuming from token
    expiration or hardcoded values, to correctly filter threads for 'this calendar month'.
- (pitfall) PITFALL (avoid): Do not rely on Python's datetime.date.today() for task dates; the
    environment may have a different 'today'. Use the task context or system-provided date if
    available.
- (When needing count of unread threads by label) Count unread threads by label 'priority-1'
    instead of cleaning read threads excluding priority-starred.
- (When counting threads filtered by a specific label (e.g., priority-2) and read status.) Used
    the `label` filter parameter on `show_inbox_threads` to directly fetch only priority-2 read
    threads, instead of fetching all read threads and then filtering by labels.
- (When counting unread threads with a specific priority label) Count unread threads filtered
    by a specific label (e.g., priority-3) instead of cleaning read threads.
- (pitfall) PITFALL (avoid): Fetched only inbox and outbox threads, missing sent and other
    non-archived threads. Must fetch all threads excluding archived (e.g., use labels INBOX,
    SENT, DRAFT etc. or entire 'all' minus archived).
- (pitfall) PITFALL (avoid): Do not rely solely on thread-level participants; use show_thread
    to inspect each email within a thread to accurately determine if any coworker is a sender
    or recipient.
- (when user wants to manage spam folder by rescuing coworker threads and purging all other
    spam) Focus on spam folder: identify coworker threads in spam, unspam and mark unread, then
    delete all remaining spam (vs. cleaning read threads except priority starred).
- (when the user wants to manage spam folder by recovering friends' threads and deleting the
    rest) Instead of cleaning read threads, this variant targets spam threads: identifies
    friends' threads in spam, unspams and marks them unread, then deletes all remaining spam.
- (when managing spam and preserving specific threads) Focus on spam folder; conditionally
    unspam/restore roommate threads, then delete all other spam.
- (pitfall) PITFALL (avoid): Check all mailboxes (inbox, outbox, archived) for TODO threads
    before replying and completing; the manager may have tasks in multiple folders.
- (When user has weekly 'To Do' emails from manager to reply based on star status) Process
    weekly 'To Do' emails from manager: reply based on starred status and unstar completed,
    preserving non-todo starred.
- (when closing weekly manager assignments (subject 'Assignment') by replying 'Finished.' or
    'Not Finished.' based on starred flag) Reply to weekly Assignment threads based on starred
    status, unstar finished, preserve non-todo starred.
- (When the task is to label Gmail threads by sender relationship (work vs personal) using
    contacts from the Phone app.) Uses Phone app contacts to categorize senders as coworkers or
    personal, then labels threads accordingly as 'work' or 'personal', instead of cleaning read
    threads. Also processes both inbox and outbox, not just read threads.
- (When labeling emails based on contact categories from phone app) Uses phone contacts to
    label threads by participant relationship (coworker vs personal).
- (when Gmail threads need to be labeled based on contact relationships from Phone app) Added
    step to retrieve contacts from Phone app, categorize by relationship
    (coworkers/friends/family), then apply 'work' or 'personal' labels to email threads based
    on sender/recipient matching those contacts.
\end{verbatim}
\rule{\linewidth}{0.7pt}
\end{minipage}
\caption{The 23 bullets the workflow reflector accumulated onto the node of Figure~\ref{fig:node-workflow} over 24 later episodes. 7 are \textsc{pitfall} bullets distilled from \emph{failed} episodes; the rest are variant bullets from \emph{solved} near-neighbour episodes. The procedure itself was never rewritten.}
\label{fig:node-bullets}
\end{figure}

\begin{figure}[t]
\centering
\begin{minipage}{\textwidth}
\rule{\linewidth}{0.7pt}
\fontsize{7.8}{9.4}\selectfont
\begin{verbatim}
---
id: meta_app_gmail_2
type: meta_framework
semantic_role: task_pattern
tags: [app:gmail, obj:attachment, obj:email, obj:email_invitations, obj:email_thread,
obj:files, op:create, op:transfer, solved]
applicability: "'app:gmail' family"
---
Common pattern generalized from: wf_send_hiring_email_replies,
wf_download_email_attachment_to_filesystem, wf_email_files_per_text_request

**Why:** Recurring community of similar workflows.

**How to apply:** Route any 'app:gmail'-like task through the matching workflow.
- (when hiring and need to email candidates) variant wf_send_hiring_email_replies: Send hiring
    email replies
- (When needing to transfer email attachment to local folder) variant
    wf_download_email_attachment_to_filesystem: Download Email Attachment to Filesystem
- (When a contact requests files via text) variant wf_email_files_per_text_request: Email files
    per text request
- (when phone message requests emailing a file) variant wf_email_file_from_phone_request: Email
    file from phone request
- (Forward email thread with extra file and prefixed note) variant
    wf_forward_thread_with_note_and_attachment: Forward thread with note and attachment
- (When you need to send personalized bulk emails using a draft template and contact list from
    file.) variant wf_send_personalized_bulk_emails_from_draft: Send personalized bulk emails
    from draft
- (when need to forward one email from a thread) variant wf_forward_single_email_from_thread:
    Forward single email from thread
- (When deleting read Gmail threads, keeping only priority starred ones) variant
    wf_clean_gmail_read_threads_except_priority_star: Clean Gmail read threads except priority
    starred
- (When away and need to reply to recent unread emails) variant
    wf_vacation_auto_reply_to_unresponded_threads: Vacation auto-reply to unresponded threads
- (when draft email contains $TODO prices for Amazon items) variant
    wf_fill_draft_prices_from_amazon: Fill draft prices from Amazon
- (When need to forward email with prefixed comment) variant wf_forward_email_with_prefix_note:
    Forward email with prefix note
- (when needing to forward email with custom prefix) variant
    wf_forward_email_with_prefixed_note: Forward email with prefixed note
\end{verbatim}
\rule{\linewidth}{0.7pt}
\end{minipage}
\caption{The \textbf{meta-framework} node that generalizes the workflow of Figure~\ref{fig:node-workflow}, induced automatically by community detection over the workflow similarity graph.}
\label{fig:node-meta}
\end{figure}

\end{document}